%% file: main.tex
\documentclass[preprint,10pt]{elsarticle}

\usepackage{amssymb}
\usepackage{hyperref}       
\usepackage{url}            
\usepackage{booktabs}       
\usepackage{multirow}
\usepackage{amsfonts}       
\usepackage{nicefrac}       
\usepackage{microtype}      
\usepackage{graphicx}
\usepackage{amsmath}

\journal{Neurocomputing}

\begin{document}

\begin{frontmatter}

\title{Dissimilarity Mixture Autoencoder for Deep Clustering}

\author[inst1]{Juan S. Lara}
\author[inst1]{Fabio A. Gonz\'alez}

\affiliation[inst1]{organization={Universidad Nacional de Colombia},
            city={Bogota},
            postcode={111321},
            country={Colombia}}

\input{abstract/abstract.tex}

\end{frontmatter}
\input{introduction/introduction.tex}
\input{related_work/related_work.tex}

\input{dmae/dmae.tex}

\input{experiments/experiments.tex}
\input{conclusions/conclusions.tex}

\bibliographystyle{elsarticle-num}
\bibliography{references.bib}

\end{document}

%% file: abstract/abstract.tex
\begin{abstract}
    The dissimilarity mixture autoencoder (DMAE) is a neural network model for feature-based clustering that incorporates a flexible dissimilarity function and can be integrated into any kind of deep learning architecture. It internally represents a dissimilarity mixture model (DMM) that extends classical methods like K-Means, Gaussian mixture models, or Bregman clustering to any convex and differentiable dissimilarity function through the reinterpretation of probabilities as neural network representations. DMAE can be integrated with deep learning architectures into end-to-end models, allowing the simultaneous estimation of the clustering and neural network's parameters. Experimental evaluation was performed on image and text clustering benchmark datasets showing that DMAE is competitive in terms of unsupervised classification accuracy and normalized mutual information. \footnote{The source code with the implementation of DMAE is publicly available at: \url{https://github.com/juselara1/dmae}}
\end{abstract}

%% file: introduction/introduction.tex
\section{Introduction}

Unsupervised learning (UL) aims to automatically extract meaningful patterns from unlabeled data, it covers different tasks like clustering, density estimation, dimensionality reduction, anomaly detection, data generation, among others. In recent years, deep learning has been used as an important approach for UL, specifically, neural networks are able to automatically learn high-level abstractions of the data through unsupervised representation learning. Some remarkable examples include: autoencoders (AE), which are among the most studied neural networks and have demonstrated to outperform conventional shallow approaches in the task of dimensionality reduction \cite{hinton2006reducing}; generative models like the Generative Adversarial Networks (GANs) \cite{goodfellow2014generative} or Variational Autoencoders (VAE) \cite{kingma2013auto}, that show impressive results in the generation of image data; and deep clustering methods, which show that a neural network with the appropriate layers and certain regularizers can outperform conventional clustering methods \cite{xie2016unsupervised}.

This paper addresses the task of clustering, its main purpose is to divide data points into groups or clusters, such that, points in the same group are similar and points in different groups are dissimilar. Clustering methods can be divided into two main categories \cite{jiang2016variational}: \textit{similarity-based clustering}, in which a similarity matrix between points must be computed, different similarity criteria allow a better approximation of the cluster's shapes or densities and includes methods like hierarchical clustering, K-Medoids, spectral clustering \cite{ng2002spectral} and affinity propagation \cite{frey2007clustering}. On the other hand, \textit{feature-based clustering} which allows estimating the clusters using a feature representation of each sample while assuming independent and identically distributed (\textit{iid}) samples, it includes methods like K-means, Gaussian mixture models (GMM), clustering with Bregman divergences \cite{banerjee2005clustering} and the recent deep clustering methods \cite{xie2016unsupervised}. 

This paper presents the dissimilarity mixture autoencoder (DMAE), a deep neural network model for feature-based clustering that incorporates a dissimilarity function that preserves the flexibility of similarity-based clustering. To this end, DMAE internally represents a mixture of exponential distributions as an encoder-decoder architecture that allows parameter estimation using gradient-based optimization. DMAE reinterprets classical probabilistic notions as neural network components, allowing easy incorporation into deep learning architectures. Our main contributions can be summarized as:

\begin{itemize}
    \item DMAE: a deep clustering model based on a differentiable and convex dissimilarity function between samples and mixture components that can be applied to deeply embedded spaces. It extends clustering methods like K-means, Gaussian mixture models, or clustering with Bregman divergences \cite{banerjee2005clustering} using a general dissimilarity function and deep representation learning.
    \item The proposed model is an encoder-decoder network that uses probabilistic neural network representations and can be used alongside other deep learning components. This allows scalable parameter estimation through gradient-based optimization, online learning, and modern strategies like transfer learning. 
    \item Due to its formulation based on dissimilarity functions, DMAE can fit data that comes from different probability distributions while preserving parameter's interpretability, some examples include circular, angular, probability spaces, among others. Likewise, if the problem demands a higher complexity, the proposed model can be extended with deep learning components, performing a trade-off between interpretability and performance.
\end{itemize}

The remainder of this document is organized as follows: Section \ref{sec:related_work} presents related work in clustering and deep clustering; Section \ref{sec:dmae} introduces DMAE and presents its probabilistic motivation; Section \ref{sec:experiments} describes the experimental evaluation and presents the results; Section \ref{sec:conclusions} shows the final remarks and the future works.

%% file: related_work/related_work.tex
\section{Related Work}\label{sec:related_work}

The performance of clustering models is highly dependent on the topology and the properties of the input data, for instance, clusters may form different geometrical shapes (e.g., circular, elliptical, angular, among others) or may be composed of different types of variables (e.g., continuous, discrete, among others). Different clustering strategies are more suitable for different problems, nevertheless, they can be formulated under two main paradigms as it will be described in the following sections.

\input{related_work/similarity_based.tex}
\input{related_work/feature_based.tex}

%% file: related_work/similarity_based.tex
\subsection{Similarity-based Clustering}

These methods rely on the selection of a similarity function that is used to compute an affinity matrix that represents how similar are different samples. One of the main advantages of similarity-based methods is their flexibility, different similarity functions can be used according to the complexity of the data. For instance, the K-Medoids \cite{park2009simple} method considers a sample-wise similarity matrix to determine a set of $K$ samples as clusters descriptors, likewise, affinity propagation \cite{frey2007clustering} performs a similar task but automatically determines the optimal number of clusters. There are other methods like spectral clustering \cite{ng2002spectral} or hierarchical clustering \cite{johnson1967hierarchical} which also require an affinity matrix to perform spectral decomposition of Laplacian matrices or to determine association rules.

Similarity-based strategies are usually the best option for small datasets since they have the flexibility to describe individual relations between samples. Nevertheless, the main problem of these methods is the computation of the affinity matrix, which has high time and memory complexities and does not scale well to larger datasets. One of the most interesting similarity-based clustering methods is kernel K-Means \cite{dhillon2004kernel}, which is a method that follows a feature-based formulation and results in a similarity-based method due to the kernel trick. Kernel K-Means has demonstrated to generalize other clustering methods like K-Means, spectral clustering, among others \cite{dhillon2004kernel,ye2008discriminative}, however, it has the same scalability issue since it requires the computation of a Gram matrix. Some alternatives have been proposed to address this issue, including the Nystr\"om approximation \cite{wang2019scalable}, random Fourier features \cite{chitta2012efficient}, randomized matrix factorization, among others. However, this is still an active research problem and has not been fully explored. 

%% file: related_work/feature_based.tex
\subsection{Feature-based Clustering}

Feature-based methods assume \textit{i.i.d} samples and perform clustering over a feature representation of each sample, therefore, inter-sample relationships are not considered. Most feature-based strategies can be represented as latent variable models, in which the problem lies in the estimation of hidden or unknown cluster parameters. The simplest case is K-Means, which aims to determine a set of $K$ latent centroids by iteratively minimizing the mean squared error between samples and the latent centroids, it is one of the most popular clustering methods due to its simplicity and its scalability, allowing distributed and parallel computation to ease the training on large datasets. The main disadvantages of K-Means are the linear behavior and the hard-assignment approach, which in most cases do not allow the capture of relevant data patterns or represent uncertainty. For this reason, more general methods have been proposed, for instance, fuzzy c-means \cite{bezdek1984fcm} extends K-Means for soft clustering, it assigns each sample to all the clusters using a weight that is computed from normalized Euclidean distances. Other methods have a probabilistic formulation, the most popular is the Gaussian Mixture Model (GMM) \cite{reynolds2009gaussian}, which soft-assigns each point to each cluster according to a posterior distribution that is computed from normal cluster likelihoods and multinomial priors. The feature-based method that is more closely related to our proposed model is Bregman Soft Clustering \cite{banerjee2005clustering}, it is a model that soft-assigns each sample through a mixture of exponential distributions and generalizes other methods when choosing different divergences, e.g., K-Means is equivalent to a Euclidean distance, GMM is equivalent to the Mahalanobis distance (GMM), among others. It is a flexible model that estimates its parameters through a procedural algorithm based on expectation-maximization and can use different similarity functions whereas they can be expressed as a Bregman divergence. Like similarity-based methods, the importance of Bregman clustering lies in the flexibility of the selection of different Bregman divergences, without requiring the computation of a sample-wise affinity matrix. 

In recent years, deep clustering methods have gained a special interest in feature-based clustering, these methods use a deep neural network (DNN) to transform the data into simpler and cluster-friendly representations \cite{aljalbout2018clustering}, it aims to induct a latent space through several non-linear transformations where clustering is performed. This is useful if the clustering strategy allows differentiation since a DNN can be fine-tuned through back-propagation to improve the clustering results. A pioneer work in deep clustering is the Deep Embedded Clustering (DEC) \cite{xie2016unsupervised}, it uses an autoencoder to transform the data into a low-dimensional space and an assignment distribution layer to refine the latent space for clustering. The original method consisted of a stacked autoencoder of fully-connected layers for the representation and a t-student distribution for the cluster assignments, however, recent studies have presented improved versions that include convolutional autoencoders \cite{guo2017deep}, boosting \cite{li2018discriminatively} and data augmentation \cite{guo2018deep}. DEC has demonstrated that deep clustering can outperform conventional shallow approaches in different tasks, this has generated a new line of research in both deep learning and feature-based clustering.

The core of current deep clustering methods is the assignment strategy, i.e., determining a specific operation to compute which cluster is more suitable for each data point while allowing backpropagation. For instance, the t-student distribution has been widely used, especially, it can be seen as a normalized Euclidean similarity that is able to represent the K-Means behavior. An alternative is a multinomial logistic regression or SoftMax function, it is typically used in deep learning and allows to approximate the one-hot assignments or the argmax function under certain constraints \cite{agustsson2017soft}. This last approach was explored in the Deep Embedded Regularized Clustering (DEPICT) \cite{ghasedi2017deep}, which incorporates additional regularization terms to allow a uniform distribution of the assignments. Similar studies in vector quantization have explored this idea in detail, for instance, the soft-to-hard vector quantization method \cite{agustsson2017soft} uses a softmax function over euclidean distances as a relaxation to the nearest neighbor assignments that are typically performed in vector quantization. Generative models have been also proposed for deep clustering. For instance, the Variational Deep Embedding (VaDE) \cite{jiang2016variational} and the Gaussian Mixture Variational Autoencoder (GMVAE) \cite{dilokthanakul2016deep} use a process that generates random samples in the deep embedded space from multivariate normal distributions. In a similar manner, adversarial-based approaches like the InfoGAN \cite{chen2016infogan} and the ClusterGAN \cite{mukherjee2019clustergan} have been successfully applied. 

Although most of the deep clustering methods are related to feature-based clustering, some works have extended similarity-based methods for deep clustering. For example, the Joint Unsupervised Learning (JULE) \cite{yang2016joint} uses an agglomerative loss that requires an affinity matrix to represent the similarities between each point in the dataset. The SpectralNet \cite{shaham2018spectralnet} is a neural network model that can learn the spectral map that is computed in spectral clustering, also, it uses a siamese network to learn a similarity function between points. Finally, deep clustering via a Gaussian mixture variational autoencoder with graph embedding (DGG) \cite{yang2019deep} is a generative model that extends VaDE, it uses a graph embedded affinity matrix that is also constructed using a siamese network. These methods combine the flexibility of similarity-based clustering and the representation learning capabilities of neural networks, nevertheless, they still require to compute or approximate an affinity or similarity matrix.

%% file: dmae/dmae.tex
\section{Dissimilarity Mixture Autoencoder}\label{sec:dmae}

The dissimilarity mixture autoencoder (DMAE) is a neural network model for feature-based clustering that incorporates a flexible dissimilarity function and can be integrated into any kind of deep learning architecture. The formulation of DMAE consists of three main components. First, it internally incorporates a \textit{dissimilarity mixture model} (DMM) that extends classical methods like Bregman clustering to any convex and differentiable dissimilarity function; second, an \textit{autoencoder for expectation-maximization} is proposed as the learning procedure for the DMM; and third, \textit{unsupervised representation learning} is used to extend the model for deep clustering.

\input{dmae/dmm.tex}
\input{dmae/ae_for_em}
\input{dmae/unsupervised_representation_learning.tex}

%% file: dmae/dmm.tex
\subsection{Dissimilarity Mixture Model}\label{sec:dmm}

Mixture models are a probabilistic approach for clustering that allows representing the density of the samples through the combination of a set of $K$ distributions. In this work, the likelihood $P(\mathbf{x}_i|z_{ik}=1)$ of a data point $\mathbf{x}_i\in {\rm I\!R^{m}}$ that belongs to a cluster $k$ is modeled as an exponential distribution as shown in Eq. \ref{eq:likelihood}.
\begin{equation}\label{eq:likelihood}
    P(\mathbf{x}_i|z_{ik}=1; \theta_k) = \text{exp}(-\alpha d(\mathbf{x}_i, \theta_k))b_k
\end{equation}
Where $z_{ik}$ is a binary latent variable that indicates if the sample $\mathbf{x}_i$ belongs to a cluster $k$, $b_k$ is a uniquely determined value that normalizes the likelihood to be a valid probability density, $\alpha$ is a constant that controls the exponential behavior of the likelihood, and $d(\cdot)$ is a dissimilarity function that measures the affinity between $\mathbf{x}_i$ and the parameters $\theta_k$ associated to the cluster $k$. There is a direct connection between exponential families and the proposed likelihood when $d(\cdot)$ is a Bregman divergence \cite{banerjee2005clustering}. Nevertheless, the proposed likelihood distribution considers a more general differentiable dissimilarity function that must be convex to $\theta_k$. Some remarkable examples of functions that can be used as dissimilarity functions are presented in Table \ref{tab:dissimilarities}.

\begin{table}[ht!]
\centering
\caption{Example of differentiable and convex functions that can be used as dissimilarities.}
\begin{tabular}{ccc}
\toprule
    Name & Expression & Parameters\\ 
    \midrule
    Euclidean & $||\mathbf{x}-\mu_k||$ & $\theta_k=\mu_k$\\
    Manhattan & $|\mathbf{x}-\mu_k|$ & $\theta_k=\mu_k$\\
    Dot product & $\mathbf{x} \cdot \mu_k$ & $\theta_k=\mu_k$\\
    Itakura-Saito & $\sum_j \frac{x_j}{y_j} - \log{\frac{x_j}{y_j}} - 1$ & $\theta_k=\mathbf{y}_k$\\
    Kullback-Leibler & $\sum_{j=1}^m x_j \log{\frac{x_j}{q_j}}$ & $\theta_k=\mathbf{q}_k$ \\
    Mahalanobis & $\sqrt{(\mathbf{x} - \mu_k)^T \Sigma_k (\mathbf{x} - \mu_k)}$ & $\theta_k=(\mu_k, \Sigma_k)$\\
    \bottomrule
\end{tabular}
\label{tab:dissimilarities}
\end{table}

Following a typical mixture model formulation, it is necessary to determine an expression for the posterior cluster assignment distribution $P(z_{ik}=1|\mathbf{x}_i)$, to this end, we consider a marginal prior distribution $P(z_{ik}=1)=\pi_k$ such that $\pi_k \in [0, 1]$ and $\sum_{i=1}^K \pi_k = 1$, where $K$ is the number of clusters. This marginal distribution represents the discrete nature of the $z_{ik}$ components and introduces the mixture coefficients $\pi_k$, which are additional model parameters. The posterior distribution can be determined through the Bayes rule as presented in Eq. \ref{eq:posterior}.
\begin{equation}\label{eq:posterior}
    P(z_{ik}=1|\mathbf{x}_i) = \frac{\text{exp}(-\alpha d(\mathbf{x}_i, \theta_k))\pi_k b_k}{\sum_{j=1}^K \text{exp}(-\alpha d(\mathbf{x}_i, \theta_j))\pi_j b_j}
\end{equation}
The factor $b_k$ can be analytically determined for Bregman divergences \cite{banerjee2005clustering}, nonetheless, this is not the case for an arbitrary dissimilarity function. For this reason, we propose a reparameterization $\phi_k=\log{(\pi_k b_k)}$ to include this factor as a parameter of the model, allowing the optimization over the non-normalized probability distributions. Likewise, the expression presented in Eq. \ref{eq:softmax} shows that the posterior distribution $P(z_{ik}=1|\mathbf{x}_i)$ or the \textit{responsibility} that component $k$ takes over $\mathbf{x}_i$ can be reinterpreted as the Gumbel-softmax distribution \cite{jang2016categorical}. This is important since the softmax $\sigma(\cdot)$ is a well-known function that is commonly used in deep learning.
\begin{equation}\label{eq:softmax}
    P(z_{ik}=1|\mathbf{x}_i) = \sigma(-\alpha d(\mathbf{x}_i, \theta_k)+\phi_k))
\end{equation}
The $\alpha$ value can be also reinterpreted as the \textit{softmax inverse temperature}, which is a hyperparameter that controls the sparcity of the softmax function, higher $\alpha$ values approximate the softmax outputs to one-hot assignments. Similarly, there is a direct connection between this expression and feed-forward neural networks when the dissimilarity function is a negative dot product $d(\mathbf{x}_i,\theta_k)= -\mathbf{x}_i\cdot \theta_k$ and $\alpha=1$, the posterior $P(\mathbf{z}_i|\mathbf{x}_i)$ for all the components $\mathbf{z_i}=[z_{i1}, z_{i2}, \dots, z_{iK}]$ would be equivalent to a single-layer network with a softmax activation function $\sigma(\mathbf{W} \cdot \mathbf{x}_i + \mathbf{b})$, where $\mathbf{W} \in {\rm I\!R^{K \times m}}$ is a matrix that contains the cluster's parameters $\theta_k$ and $\mathbf{b} \in {\rm I\!R^{K}}$ is a vector with the reparameterized mixing coefficients $\phi_k$.

Although the model's parameters $\Theta=\{\theta_1, \theta_2, \dots, \theta_K\}$ and $\Phi=\{\phi_1, \phi_2, \dots, \phi_K\}$ can be estimated through maximum likelihood estimation, a direct optimization of the log-likelihood for a mixture model is not a well-posed problem and suffers from different issues like sample or posterior collapse and identifiability \cite{bishop2006pattern}. Similarly, an analytical expectation-maximization (EM) procedure has been proposed for Bregman divergences \cite{banerjee2005clustering}, nevertheless, there is not an exact procedure for every dissimilarity function. For this reason, the learning procedure of DMAE is formulated as a reinterpretation of \textit{E-step} and the \textit{M-step} as the encoding, decoding, and optimization stages of an autoencoder.

%% file: dmae/ae_for_em.tex
\subsection{Autoencoder for Expectation-Maximization}\label{sec:ae_em}

EM is an algorithm that iteratively updates the model's parameters until convergence, it requires to determine general expressions for the \textit{E-step} and the \textit{M-step} at any $t$ iteration. As shown in Eq. \ref{eq:e_step}, the \textit{E-step} consists on the computation of a proposal distribution $q^{(t)}$. For models with a discrete number of latent variables the proposal distribution is equal to the posterior distribution $P(\mathbf{Z}|\mathbf{X}; \Theta^{(t)}, \Phi^{(t)})$ that was derived in Eq. \ref{eq:posterior}. Where $\mathbf{X}\in {\rm I\!R^{N \times m}}$ is a matrix in which each row is a feature vector $\mathbf{x}_i$ of size $m$ for the $N$ samples, and $\mathbf{Z} \in {\rm I\!R^{N \times K}}$ is a matrix in which each row represents the latent components $\mathbf{z}_i$ for a sample $\mathbf{x}_i$. 
\begin{equation}\label{eq:e_step}
    q^{(t)} = P(\mathbf{Z}|\mathbf{X}; \Theta^{(t)}, \Phi^{(t)})
\end{equation}
 In the \textit{M-step} the parameters are updated by maximizing the expected value of the complete log-likelihood under the distribution $q^{(t)}$ as presented in Eq. \ref{eq:m_step}.
\begin{equation}\label{eq:m_step}
    \Theta^{(t + 1)}, \Phi^{(t + 1)} = \underset{\Theta, \Phi}{\text{argmax}} \quad {\rm I\!E}_{q^{(t)}}[\log{P(\mathbf{X}, \mathbf{Z}; \Theta, \Phi )}]
\end{equation}
The complete likelihood $P(\mathbf{X}, \mathbf{Z}; \Theta, \Phi)$ of the DMM for a set of $N$ samples and with $K$ components takes the form of Eq. \ref{eq:complete_likelihood}. 
\begin{equation}\label{eq:complete_likelihood}
    P(\mathbf{X}, \mathbf{Z}; \Theta, \Phi) = \prod_{i=1}^N \prod_{k=1}^K \pi_k^{z_{ik}} \left(\text{exp}(-\alpha d(\mathbf{x}_i, \theta_k))b_k\right)^{z_{ik}}
\end{equation}
Therefore, the complete log-likelihood would be:
\begin{equation}\label{eq:complete_log_likelihood}
    \log{(P(\mathbf{X}, \mathbf{Z}; \Theta, \Phi)}) = \sum_{i=1}^N \sum_{k=1}^K z_{ik} (\phi_k -\alpha d(\mathbf{x}_i, \theta_k))
\end{equation}
Considering that the $z_{ik}$ components are binary random variables and the definition of $q^{(t)}$ presented in Eq. \ref{eq:e_step}, the conditional expectation ${\rm I\!E}_{q^{(t)}}[z_{ik}]$  is equal to the posterior distribution $P(z_{ik}=1|\mathbf{x}_i)$. Hence, the expectation $L=\mathbb{E}_{q^{(t)}}[\log{(P(\mathbf{X}, \mathbf{Z}; \Theta, \Phi)})]$ of the complete log-likelihood takes the form of Eq. \ref{eq:complete_log_likelihood2}. 
\begin{equation}\label{eq:complete_log_likelihood2}
    L = \sum_{i=1}^N \sum_{k=1}^K P(z_{ik}=1|\mathbf{x}_i) (\phi_k -\alpha d(\mathbf{x}_i, \theta_k))
\end{equation}
Since the dissimilarity function must be convex, we can formulate a Jensen's inequality as shown in Eq. \ref{eq:jensen}.
\begin{equation}\label{eq:jensen}
    \sum_{k=1}^K P(z_{ik}=1|\mathbf{x}_i)d(\mathbf{x}_i, \theta_k) \geq d\left(\mathbf{x}_i, \sum_{k=1}^K P(z_{ik}=1|\mathbf{x}_i)\theta_k\right)
\end{equation}
Using the inequality of Eq. \ref{eq:jensen} and considering that the complete log-likelihood must be maximized, we can determine a lower bound as shown in Eq. \ref{eq:lower_bound}.
\begin{equation}\label{eq:lower_bound}
    L \geq \sum_{i=1}^N \left( \widetilde{\phi}_i - \alpha d(\mathbf{x}_i, \widetilde{\theta}_k) \right)
\end{equation}
Where the parameters $\widetilde{\theta}_i$ and $\widetilde{\phi}_i$ for a sample $\mathbf{x}_i$ are a convex combination of all the cluster's parameters using the softmax outputs as weights.
\begin{equation}\label{eq:assigned_params}
\begin{split}
    \widetilde{\theta}_i = \sum_{k=1}^K P(z_{ik}=1|\mathbf{x}_i) \theta_k\\
    \widetilde{\phi}_i = \sum_{k=1}^K P(z_{ik}=1|\mathbf{x}_i) \phi_k
\end{split}
\end{equation}
The loss function $\mathcal{L}$ to minimize is shown in Eq. \ref{eq:dmm_loss}, it is based on the lower bound of the complete log-likelihood and the soft-assigned parameters, and is similar to the distortion measure that is typically used in other clustering methods.
\begin{equation}\label{eq:dmm_loss}
    \mathcal{L} = \sum_{i=1}^N \alpha d\left(\mathbf{x}_i, \widetilde{\theta}_i\right) - \widetilde{\phi}_i
\end{equation}
The dissimilarity mixture autoencoder (DMAE) is the reinterpretation of EM procedure as an encoder-decoder network with shared weights that correspond to the DMM's parameters. In DMAE, an encoding process is performed to compute the soft-assignments or the posterior probability $P(\mathbf{z}_i|\mathbf{x}_i)$, which is equivalent to the \textit{E-step}; likewise, reconstructions $\widetilde{\theta}_i$ and $\widetilde{\phi}_i$ are computed as a linear decoding of the latent representation, and a reconstruction error or distortion measure $\mathcal{L}$ is minimized to update the model's parameters, which is equivalent to the \textit{M-step}.

One of the most important properties of DMAE, is that it is composed of differentiable operations, therefore, $\mathcal{L}$ can be minimized through gradient-based optimization. Approaches like batch learning and online learning can be used since the DMM assumes i.i.d. samples, allowing its application to large datasets with controlled memory consumption. In addition, DMAE can be enhanced using unsupervised representation learning, by exploiting back-propagation and the external gradients of other deep learning components.

%% file: dmae/unsupervised_representation_learning.tex
\subsection{Unsupervised Representation Learning}\label{sub:unsupervised_representation_learning}

\begin{figure}[t!]
    \centering
    \includegraphics[width=1.0\columnwidth]{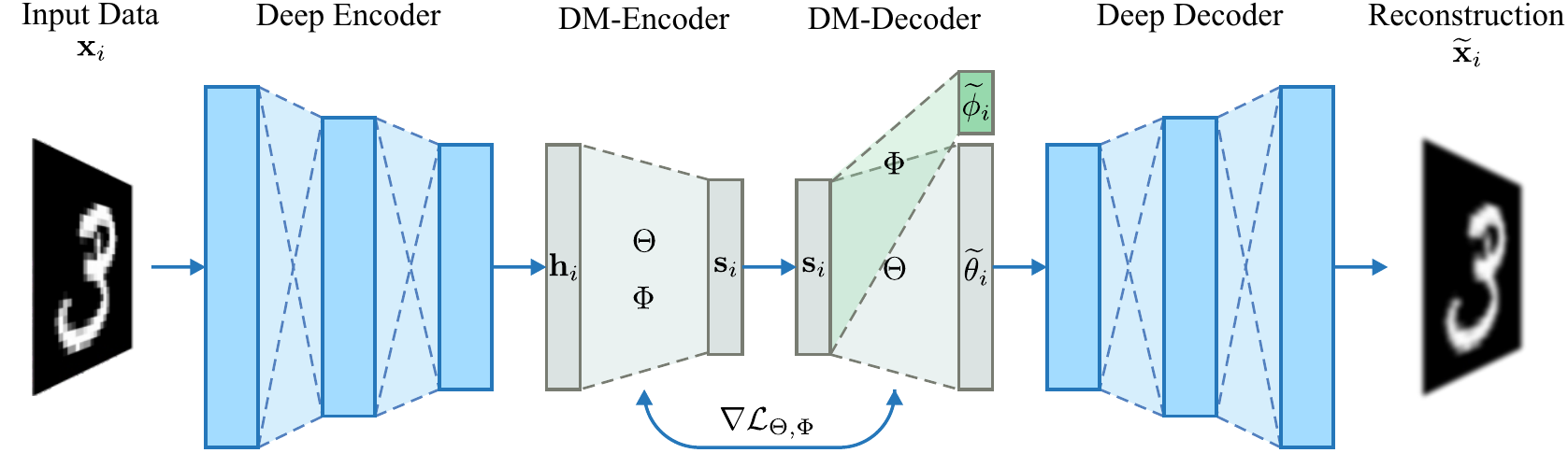}
    \caption{Conceptual diagram of the dissimilarity mixture autoencoder.}
    \label{fig:dmae}
\end{figure}

A conceptual diagram of DMAE for deep clustering is depicted in Fig. \ref{fig:dmae}, it uses a deep autoencoder for unsupervised representation learning and incorporates the DMM in the latent space for clustering. The complete architecture can be divided into four main components: \textit{deep encoder}, \textit{dissimilarity mixture encoder}, \textit{dissimilarity mixture decoder} and \textit{deep decoder}.

\subsubsection{Deep Encoder}

The deep encoder component transforms the input data into a simpler representation or latent space. It defines a mapping $\mathcal{X} \rightarrow \mathcal{H}$ from the original input space $\mathcal{X}$ to a space $\mathcal{H}$ of dimension $R$. This is achieved through several non-linear transformations that must be learnt during training. Specifically, it defines a function $f_1(\cdot)$ that transforms an input sample $\mathbf{x}_i \in {\rm I\!R^{m }}$ to a latent representation $\mathbf{h}_i \in {\rm I\!R^{R}}$ ($R$ is the size of the latent dimension) using a set of weights $W_e$.
\begin{equation}\label{eq:deep_encoder}
    \mathbf{h}_i=f_1(\mathbf{x}_i, W_e)
\end{equation}
\subsubsection{Dissimilarity Mixture Encoder (DM-Encoder)}

This component soft-assigns a latent representation $\mathbf{h}_i$ into the $K$ clusters. It defines a mapping $\mathcal{H} \rightarrow \mathcal{S}$ from the space $\mathcal{H}$ to a latent space $\mathcal{S}$ of dimension $K$. This representation is equivalent to the assignments $P(z_k=1|\mathbf{x}_i)$ of the DMM that are shown in Eq. \ref{eq:posterior} and varies between a uniformly distributed ($\alpha \rightarrow 0$) and a sparse space ($\alpha \rightarrow \infty$). It defines the clusters parameters $\Theta, \Phi$ and requires a pairwise dissimilarity $d_p(\cdot)$ to compute a vector of dissimilarities $\mathbf{d}_i \in {\rm I\!R^{K }}$, such that each value in $\mathbf{d}_i$ represents the dissimilarity $d(\cdot)$ between $\mathbf{h}_i$ and all the cluster parameters $\theta_k \in \Theta$ as presented in Eq. \ref{eq:dissimilarity}.
\begin{equation}\label{eq:dissimilarity}
   \mathbf{d}_i = d_p(\mathbf{h}_i, \Theta)
\end{equation}
The soft-assignments $\mathbf{s}_i\in {\rm I\!R^{K }}$ are hence determined through the softmax activation function $\sigma(\cdot)$, it uses the \textit{softmax inverse temperature} $\alpha$ to control the sparsity and a vector of biases (reparameterized mixing coefficients) $\Phi \in {\rm I\!R^{K }}$ as shown in Eq. \ref{eq:soft_assignments}.
\begin{equation}\label{eq:soft_assignments}
\centering
    \mathbf{s}_i=\sigma(-\alpha \mathbf{d}_i+\Phi)
\end{equation}

\subsubsection{Dissimilarity Mixture Decoder (DM-Decoder)}

The purpose of this component is to compute the assigned cluster's parameters $\widetilde{\theta}_i$ and $\widetilde{\phi}_i$. It defines a mapping $\mathcal{S} \rightarrow \widetilde{\Theta}$ from the sparse space $\mathcal{S}$ into a space $\widetilde{\Theta}$ of the reconstructed parameters. This can be achieved through a convex combination of all the cluster's parameters $\Theta, \Phi$ as it was shown in Eq. \ref{eq:assigned_params}. Likewise, if the parameters $\Theta$ can be structured as a matrix $\Theta \in {\rm I\!R^{K \times m}}$, then the reconstruction would be equivalent to the matrix multiplications that are presented in Eq. \ref{eq:reconstruction}, i.e., each cluster is represented by a single vector $\theta_k$ and a mixing coefficient $\phi_k$.
\begin{equation}\label{eq:reconstruction}
\centering
\begin{split}
    \widetilde{\phi}_i =  \Phi^T \cdot \mathbf{s}_i\\
    \widetilde{\theta}_i = \Theta^T \cdot \mathbf{s}_i
\end{split}
\end{equation}

\subsubsection{Deep Decoder}

This component defines a mapping $\widetilde{\Theta} \rightarrow \widetilde{\mathcal{X}}$ from the reconstructed parameter space $\widetilde{\Theta}$ to a global reconstructed space $\widetilde{\mathcal{X}}$. It defines a function $f_2(\cdot)$ that uses a set of weights $W_d$ to transform $\widetilde{\theta}_i$ into a global reconstruction $\widetilde{\mathbf{x}}_i$ in the original input space.
\begin{equation}\label{eq:deep_decoder}
\centering
    \widetilde{\mathbf{x}}_i = f_2(\widetilde{\mathbf{h}}_i, W_d)
\end{equation}
The complete model can be optimized using a composed loss function as presented in Eq. \ref{eq:deep_loss}. It is a linear combination of two terms: (1) the reconstruction loss $\mathcal{L}_r$, which is a standard loss for autoencoders and consists of the mean squared error between the input sample $\mathbf{x}_i$ and its reconstruction $\widetilde{\mathbf{x}_i}$; and (2) the clustering regularization $\mathcal{L}_c$, which is the DMM loss adapted to the latent space $\mathbf{h}_i$ as it was derived in the section \ref{sec:ae_em}.
\begin{equation}\label{eq:deep_loss}
\begin{split}
    \mathcal{L} &= \lambda_r \mathcal{L}_r + \lambda_c\mathcal{L}_c\\
    \mathcal{L}_r &= \left(\sum_{i=1}^N || \mathbf{x}_i-\widetilde{\mathbf{x}}_i ||^2\right)\\
    \mathcal{L}_c &= \left (\sum_{i=1}^N d(\mathbf{h}_i, \widetilde{\theta}_i) - \widetilde{\phi}_i \right)
\end{split}   	
\end{equation}

%% file: experiments/experiments.tex
\section{Experiments}\label{sec:experiments}

\input{experiments/synthetic.tex}
\input{experiments/real.tex}

%% file: experiments/synthetic.tex
\subsection{Synthetic data}

One of the main advantages of DMAE is that it is a flexible and interpretable model that can be used in several applications and does not depend on additional neural network components. More precisely, different dissimilarity functions are usually enough for a large number of problems, and deep representations can be used to extend the model if the problem complexity requires it. To assess this behavior, some experiments are performed over synthetic benchmark datasets with different geometrical properties, these datasets ease the visualization of clustering regions and allow a visual comparison with conventional clustering benchmark methods. 

\subsubsection{Datasets}

\begin{itemize}
    \item \textbf{Painwheel}: it's a dataset that was originally proposed by Johnson et al. \cite{johnson2016composing} and was used to validate the Gaussian mixture variational autoencoder \cite{dilokthanakul2016deep}. It can be generated from the arcs of 5 circles such that each point is obtained from distorted normal distributions with: angular standard deviation of $0.05$, a radial standard deviation of $0.3$, and a distortion rate of $0.25$.
    \item \textbf{Toroidal}: this dataset corresponds to a set of 4 Gaussian blobs with constant variances ($\sigma=0.05$) on the surface of a three-dimensional torous. The points are generated in ${\rm I\!R^{2}}$ and acceptance-rejection sampling is used to transform the points that lie outside the circular boundaries.
    \item \textbf{Moons}: it contains samples from two interleaving half circles. The samples have Gaussian noise with a standard deviation of $0.1$.
    \item \textbf{Circles}: this dataset contains points that are randomly generated around two centered circles with different sizes. The points have a standard deviation of $0.1$, and there is a factor of $0.1$ between the radius of the inner and outer circles.
\end{itemize}

\subsubsection{Experimental settings}

A total of $1000$ samples are generated for each dataset. DMAE is trained to find a number $K$ of clusters that corresponds to the number of groups defined by each generative process. Different dissimilarity functions and autoencoder architectures are used for each problem, especially: the Mahalanobis dissimilarity is used for the \textit{five arcs} data; a pairwise Euclidean distance that considers boundary constrains is used for the \textit{toroidal}; and a stacked autoencoder of dimensions $2-256-256-100-256-256-2$ is trained to transform the \textit{circles} and \textit{moons} data before the training of an end-to-end model (view Figure \ref{fig:dmae}) with DMAE in the layer of $100$ dimensions. The $\alpha$ parameter is explored between $\alpha \in [0.5, 1000]$, the learning rate is explored in a logarithm scale $lr \in [10^{-5}, 10^{-3}]$, the batch size is $32$ in all the cases and the number of epochs is explored between $[40, 250]$ for both pretraining (if applies) and training. A point is assigned to a cluster according to the rule presented in Eq. \ref{eq:prediction}, which considers the highest posterior probability or responsibility as presented in Eq. \ref{eq:posterior}.
\begin{equation}\label{eq:prediction}
    \hat{y}_i = \underset{k}{\text{argmax}}(P(z_{ik}| \mathbf{x}_i))
\end{equation}
The proposed model is compared against conventional feature based-methods like KMeans; Gaussian mixture model (GMM); and clustering with Bregman divergences (BC) using the Euclidean, Mahalanobis, squared Manhattan, and Kullback-Leibler divergences. Also, a similarity-based clustering method like spectral clustering (SC) is used due to its capability to solve problems with different geometrical properties, in this regard, the affinity matrix is computed using the nearest neighbors approach that uses the dissimilarity function of DMAE as the affinity score between samples. For the experiments, each model is trained 10 times with different initial parameters and the performance is mainly evaluated using the mean and standard deviation of two metrics. First, the \textit{unsupervised classification accuracy} (ACC) that is shown in Eq. \ref{eq:uacc}, which is a metric that has support in $[0, 1]$, such that higher values represent a better classification. This metric is defined as the number of cluster assignments or predictions $\hat{\mathbf{y}}$ that better describe the ground truth $\mathbf{y}$, and is calculated using a linear mapping $g(\cdot)$ that determines the best match between the unsupervised predictions and the original labels, and an indicator function ${\rm I\!I}(\cdot)$ that counts the number of matches over the $N$ samples.
\begin{equation}\label{eq:uacc}
    ACC = \frac{1}{N} \sum_{i=1}^N {\rm I\!I}(y_i = g(\hat{y}_i))
\end{equation}
Second, the \textit{normalized mutual information} (NMI) is a metric that ranges between $[0, 1]$, where higher values represent a higher correlation between two categorical variables. It is a measure of the mutual information $I(\cdot)$ between the assignments $\hat{\mathbf{y}}$ and the ground truth $\mathbf{y}$ and is normalized using the entropy $H(\cdot)$ of each variable as in Eq. \ref{eq:nmi}.
\begin{equation}\label{eq:nmi}
    NMI = \frac{2 I(\mathbf{y}, \hat{\mathbf{y}})}{H(\mathbf{y}) + H(\hat{\mathbf{y}})}
\end{equation}

\subsubsection{Analysis and results}

\begin{figure}[t!]
    \centering
    \includegraphics[width=1.0\columnwidth]{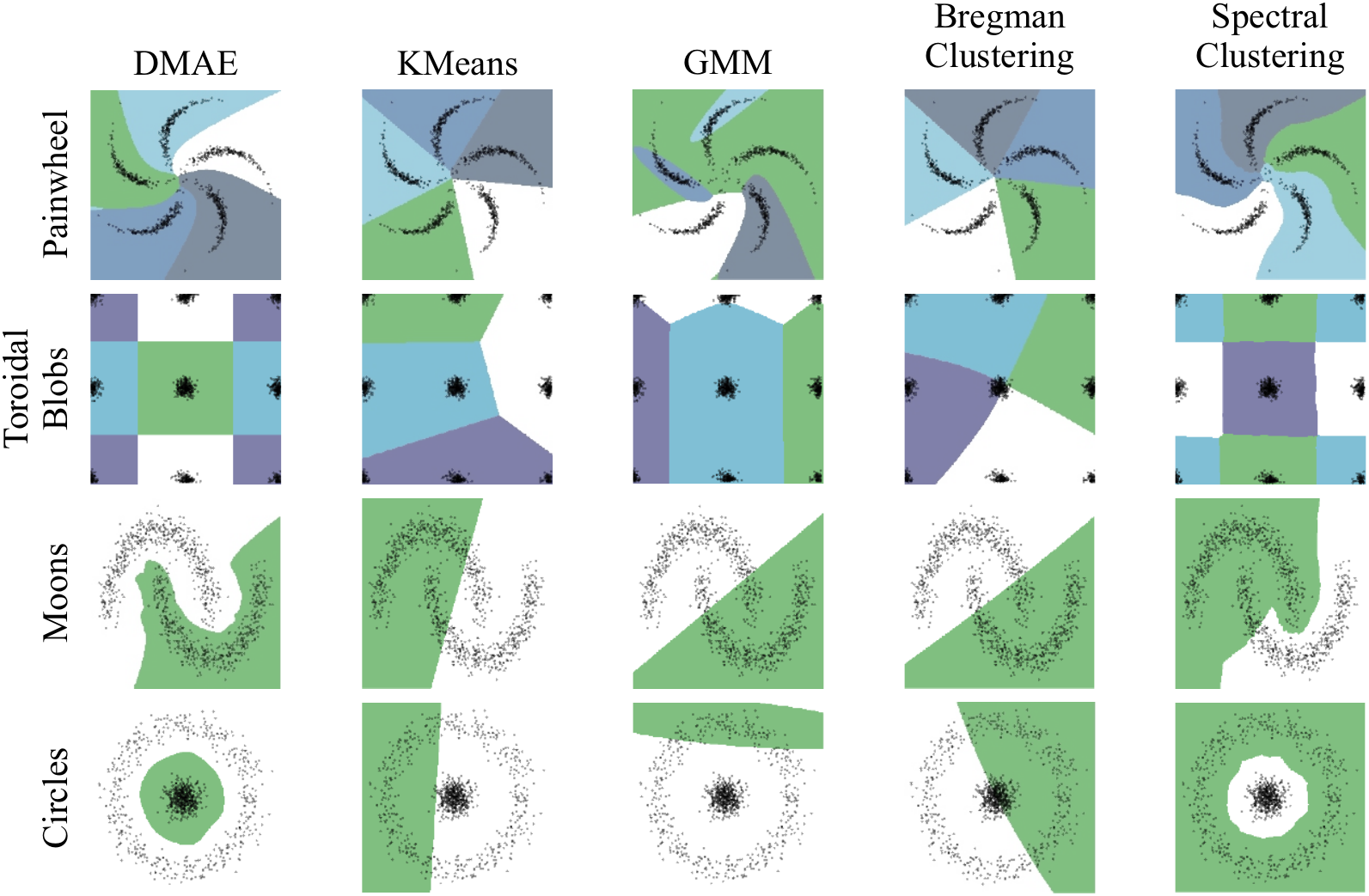}
    \caption{Clustering regions for DMAE and benchmark methods over synthetic datasets.}
    \label{fig:synthetic_results}
\end{figure}

The results are presented in Table \ref{tab:synthetic} and a visual comparison between DMAE and the baseline methods is presented in Fig. \ref{fig:synthetic_results}. These results show that DMAE can successfully capture the geometrical properties in all the cases, whereas the other feature-based benchmark methods can describe specific patterns that may not be suitable for all the datasets. Especially, a linear method like KMeans has an acceptable behavior on the \textit{painwheel} dataset, although a linear decision boundary is not the best for this case. DMAE presents a good fit using the Mahalanobis distance, while equivalent models like the GMM and BC with a Mahalanobis divergence do not present a similar result. This is a phenomenon that was detailed by Johnson et al. \cite{johnson2016composing} and is a problem that occurs with conventional density-based methods. Likewise, feature-based benchmarks are not appropriate for the \textit{toroidal blobs} dataset since these models assume distances or divergences that are unsuitable for circular data. DMAE can solve this problem because it uses a general dissimilarity function that is only constrained to be convex, therefore, it can easily consider circular boundaries without requiring a rigorous divergence formulation.

\begin{table}[ht!]
\centering
\caption{Average over 10 trials of the unsupervised classification accuracy and the normalized mutual information for the synthetic datasets.}
\begin{tabular}{@{}ccccccccc@{}}
\toprule
\multirow{2}{*}{Method} & \multicolumn{2}{c}{Painwheel} & \multicolumn{2}{c}{Toroidal} & \multicolumn{2}{c}{Moons} & \multicolumn{2}{c}{Circles}\\
 & ACC & NMI & ACC & NMI & ACC & NMI & ACC & NMI\\ \midrule
 KMeans & 0.982 & 0.943 & 0.600 & 0.495 & 0.751 & 0.190 & 0.703 & 0.245\\
 GMM & 0.908 & 0.834 & 0.585 & 0.534 & 0.850 & 0.409 & 0.618 & 0.169\\
 BC & 0.908 & 0.825 & 0.565 & 0.506 & 0.860 & 0.415 & 0.625 & 0.201\\ 
 SC & \textbf{0.985} & 0.951 & \textbf{1.000} & \textbf{1.000} & 0.878 & 0.560 & \textbf{1.000} & \textbf{1.000}\\
 \midrule
 \textbf{DMAE} & \textbf{0.985} & \textbf{0.964} & \textbf{1.000} & \textbf{1.000} & \textbf{0.949} & \textbf{0.921} & \textbf{1.000} & \textbf{1.000}\\
\bottomrule
\end{tabular}
\label{tab:synthetic}
\end{table}

Feature-based benchmark methods do not behave well in the \textit{moons} and \textit{circles} datasets, this occurs because these cases are non-linearly separable and shallow latent variable models with typical divergences are insufficient. In contrast, a similarity-based method like SC achieves higher overall performances in comparison with feature-based methods in exchange for higher computational and memory complexities, this poses a problem since it does not scale to larger datasets. The results also show that SC is sensitive to outliers, especially if noisy points connect different clusters as it occurs with the \textit{moons} case. DMAE can cluster both cases using deep representation learning, it can automatically learn an appropriate feature map for clustering and does not require the expensive computation of an affinity matrix since it only requires feature representations as input. DMAE presents a trade-off between interpretability and performance, therefore, it can deal with non-linear cases but the learned cluster's parameters lie in a deep embedded space that may not have a human-readable meaning.

%% file: experiments/real.tex
\subsection{Real data}

Unsupervised representation learning can be used to extend DMAE for deep clustering, especially, a deep representation may be more suitable for real-world problems that can not be easily modeled with a dissimilarity function.   This approach allows DMAE to cluster over deep representations that may contain more meaningful, compact, and simpler patterns. To assess this, experiments are proposed to validate the relationship between the learned clusters and human-assigned labels on image and text tasks, i.e., instead of fitting a model to predict humans' labels, the problem is addressed from an unsupervised perspective and the neural network has to learn the concepts or patterns with the minimal human intervention. 

\subsubsection{Datasets}

\begin{itemize}
    \item \textbf{MNIST}: it is comprised 28x28 images of hand-written digits in gray-scale with intensities between $[0, 255]$. It contains a total of 70.000 samples that are equally divided into 10 different categories that correspond to all the possible numbers in the positional numeral system.
    \item \textbf{REUTERS}: it contains around 810.000 English news with labels according to four categories: market, economics, corporate/industrial, and government/social. As in similar studies, the 2.000 most common terms from a sub-sample of 10.000 examples are randomly selected, and each document is represented through TF-IDF. 
    \item \textbf{USPS}: this is a dataset of digits that were scanned from envelopes by the U.S. Postal Service. It contains a total of 9.298 gray-scale images with a resolution of 16x16; It has 10 different categories corresponding to the numbers 0-9, the images are centered and contain several font styles.
\end{itemize}

\subsubsection{Experimental settings}

Similar to related studies in deep clustering \cite{aljalbout2018clustering}, We take a fine-tuning approach that consists of three steps: (1) the deep autoencoder is trained to learn a suitable feature map, (2) an embedded representation is computed using the deep encoder and a shallow clustering strategy is used on the latent space to find a set of initial cluster parameters for DMAE, (3) the end-to-end deep clustering model is trained. We consider an architecture that has been used in \cite{xie2016unsupervised, guo2017deep, guo2018deep} as the deep autoencoder, which is a stacked autoencoder of fully-connected layers (ReLu activation function for all layers except for the latent and the output spaces, which are linear) that defines an encoder of dimensions $D$-500-500-2000-$L$ and a decoder of dimensions $L$-2000-500-500-$D$, where $D$ is the size of the input space (784 for MNIST, 2000 for REUTERS and 256 for USPS) and the dimension of the latent space of the deep autoencoder $L$ is explored between $[5, 30]$.

An ablation study is performed to assess the training steps of DMAE, its main purpose is to evaluate different versions of DMAE without deep representation learning and unsupervised fine-tuning. First, a shallow version of DMAE that implements the dissimilarity mixture model and does not contain any transformation layers (i.e., clustering is performed directly on the original input space with different dissimilarity functions), this model is referred to as DMM. Second, a vanilla autoencoder learns offline, a deep embedded space and then the DMM is trained over static representations that are extracted from the deep autoencoder, it is referred to as AE+DMM. Third, the complete end-to-end DMAE model that simultaneously updates the representation and clustering layer.

For the MNIST and USPS datasets, data augmentation (random rotations using random degrees in the range $[-10, 10]$ and random horizontal and vertical shifts of 10\% of the total images' height and width) is used to train the deep autoencoder. For DMAE, the $\alpha$ hyperparameter is explored in the range $[1, 10000]$ and different dissimilarity functions are explored including Kullback-Leibler divergence, Mahalanobis distance, Euclidean distance, and Manhattan distance. We use K-means to initialize the model with the Euclidean dissimilarity and a shallow version of DMAE is trained on latent representations for the remaining dissimilarities. The number of epochs for pretraining and fine-tuning is explored in the range $[50, 500]$, the optimization is performed using the ADAM optimization method with a learning rate that is explored in the interval $[10^{-5}, 1]$, and the weights for the loss function $\lambda_r$ and $\lambda_c$ are explored in $[0, 1]$. 

An appropriate combination of hyperparameters is determined via random search due to the large number of possible combinations. The performance is evaluated in terms of the \textit{unsupervised classification accuracy} and the \textit{normalized mutual information} (view Eq. \ref{eq:uacc}-\ref{eq:nmi}), these metrics are reported as the average over 5 different trials with different random parameter initialization for the deep autoencoder and DMAE. Relevant state-of-the-art deep clustering methods that had been already validated in the same datasets using similar experimental settings are used for comparison, including: deep embedded clustering (DEC) \cite{xie2016unsupervised, guo2018deep}, semi-supervised deep embedded clustering (SDEC) \cite{ren2019semi}, variational deep embedding (VaDE) \cite{jiang2016variational}, joint unsupervised learning (JULE) \cite{yang2016joint}, deep clustering network (DCN) \cite{yang2017towards}, SpectralNet \cite{shaham2018spectralnet}, deep embedded regularized clustering (DEPICT) \cite{ghasedi2017deep}, deep density-based image clustering (DDC) \cite{ren2020deep}, convolutional deep embedded clustering \cite{guo2018deep}, deep clustering via a Gaussian mixture variational autoencoder with graph embedding (DGG) \cite{yang2019deep}, ClusterGAN \cite{mukherjee2019clustergan} and deep embedding clustering framework based on contractive autoencoder (DECCA) \cite{diallo2021}.

\subsubsection{Analysis and results}
 
Table \ref{tab:real} presents the results on the real data. DMAE achieves competitive performances with absolute differences in the ACC lower than 2\% regarding the best method on each case. The best performance is achieved when using the Mahalanobis dissimilarity on MNIST and USPS and the Euclidean dissimilarity on REUTERS. In general, there is not a single method with the best performance on all the datasets, some models have the highest performance in a specific dataset but are not competitive in the others. Likewise, other models like DEPICT, DDC, and ConvDEC are intended for image data only and are not applied to the REUTERS dataset. Although DMAE has not the best performance on every dataset, the results are consistent and both the NMI and the ACC are close to the ones of the best performing methods.

Regarding the ablation study, the DMM achieves performances that are equivalent to other shallow methods like KMeans. Although the DMM is more flexible thanks to the dissimilarity function, the formulation of a function that describes the complex patterns in images and texts is not straightforward. Deep models are more suitable in these cases considering their high representation capabilities, especially, deep embeddings ease the training process through the induction of clustering-friendly spaces. The results achieved with AE+DMM demonstrate this, the performance is highly improved when the DMM is trained with feature representations that are extracted from a deep autoencoder. Lastly, the end-to-end model outperforms both the shallow DMM and AE+DMM, this demonstrates the compatibility that DMAE has with other deep learning components, allowing to simultaneously improve the representations and the overall clustering results. 

\begin{table}[t!]\label{tab:real}
\centering
\caption{Average over 5 trials of the unsupervised classification accuracy and the normalized mutual information for the real datasets. Cases with - corrospond to methods that were proposed for images only.}
\begin{tabular}{@{}ccccccc@{}}
\toprule
\multirow{2}{*}{Method} & \multicolumn{2}{c}{MNIST} & \multicolumn{2}{c}{Reuters} & \multicolumn{2}{c}{USPS}\\
& ACC & NMI & ACC & NMI & ACC & NMI \\ \midrule
KMeans & 0.535 & 0.500 & 0.550 & 0.359 &  0.668 & 0.627 \\
DEC \cite{xie2016unsupervised, guo2018deep} & 0.849 & 0.816 & 0.777 & 0.571 & 0.758 & 0.769 \\
SDEC \cite{ren2019semi} & 0.861 & 0.829 & 0.679 & 0.509 & 0.764 & 0.777\\
VaDE \cite{jiang2016variational} & 0.945 & 0.876 & 0.792 & 0.722 & 0.566 & 0.512\\
JULE \cite{yang2016joint} & 0.906 & 0.872 & 0.751 & 0.774 & 0.914 & 0.881\\
DCN \cite{yang2017towards} & 0.830 & 0.813 & 0.792 & 0.814 & 0.778 & 0.853\\
SpectralNet \cite{shaham2018spectralnet}& 0.971 & 0.924 & 0.803 & 0.532 & 0.825 & 0.804\\
DEPICT \cite{ghasedi2017deep} & 0.965 & 0.917 & - & - & 0.964 & 0.927\\
DDC \cite{ren2020deep} & 0.969 & 0.941 & - & - & \textbf{0.977} & 0.939\\
ConvDEC \cite{guo2018deep} & \textbf{0.985} & \textbf{0.960} & - & - & 0.970 & \textbf{0.953}\\
DGG \cite{yang2019deep} & 0.976 & 0.885 & 0.823 & 0.820 & 0.904 & 0.820\\
ClusterGAN \cite{mukherjee2019clustergan} & 0.964 & 0.921 & 0.816 & \textbf{0.824} & 0.970 & 0.930\\
DECCA \cite{diallo2021} & 0.964 & 0.907 & \textbf{0.839} & 0.6343 & 0.773 & 0.805\\
\midrule
DMM & 0.540 & 0.501 & 0.550 & 0.402 & 0.659 & 0.611 \\
AE+DMM & 0.960 & 0.906 & 0.705 & 0.400 & 0.937 & 0.863 \\
\textbf{DMAE} & \textbf{0.984} & \textbf{0.945} & \textbf{0.827} & \textbf{0.580} & \textbf{0.965} & \textbf{0.952}\\
\bottomrule
\end{tabular}
\end{table}

The experimental evaluation showed that data augmentation has an important role when clustering image datasets. This phenomenon was evaluated for DEC \cite{guo2018deep} and also occurs with DMAE. Similarly, the hyperparameter that controls the assignments' sparsity $\alpha$ does not have any meaningful impact in the evaluated metrics, this is due to the assignment strategy presented in Eq. \ref{eq:prediction} which corresponds to a hard assignment of the most likely cluster, nonetheless, this parameter can be useful for uncertainty modeling because it allows variations between uniform and one-hot assignments.

%% file: conclusions/conclusions.tex
\section{Conclusions and future work}\label{sec:conclusions}

This paper presented the Dissimilarity Mixture Autoencoder (DMAE), a model that reinterprets a mixture model and its optimization through expectation-maximization as a neural network. The experimental evaluation showed that DMAE is very flexible and competitive, allowing the generalization of different mixture distributions according to the selected dissimilarity function. As a deep clustering model, DMAE poses an important implication that rises for most of the deep clustering methods, it is a different perspective for machine learning in which instead of fitting a model to mimic humans' annotations or labels, the model has to learn concepts with the minimal human intervention in a completely unsupervised fashion.

Current approaches are evaluated on controlled and well-known datasets, however, the application of deep clustering in real-life scenarios has not been completely explored. This is important considering the simplicity of the concepts that are in these sorts of problems, the performance on more challenging tasks that require higher human abstractions is still unknown. Future work is aimed to determine manners to reduce the dependencies to the deep autoencoder structure and methods based on metric learning to automatically find an appropriate dissimilarity function for each problem without the need for the extensive exploration of the dissimilarity function.